\pdfoutput=1

\documentclass[11pt,a4paper]{article}
\usepackage[hyperref]{acl2021}
\usepackage{times}
\usepackage{latexsym}

\usepackage{microtype}

\usepackage{hyperref}
\usepackage{comment}

\aclfinalcopy

\usepackage[T1]{fontenc}

\usepackage[utf8]{inputenc}

\usepackage{enumitem}
\usepackage{times}
\usepackage{latexsym}
\usepackage{graphicx}
\usepackage{url}
\usepackage{amsmath}
\usepackage{float}
\usepackage{xcolor}
\usepackage{subcaption}
\usepackage{multirow}
\usepackage{amsfonts}
\usepackage{booktabs}
\usepackage{bbm}
\usepackage{pifont}
\usepackage{makecell,rotating}
    \setcellgapes{5pt}

\usepackage{cleveref}

\hypersetup{colorlinks,urlcolor={blue}}

\usepackage{todonotes}
\makeatletter
\newcommand*\iftodonotes{\if@todonotes@disabled\expandafter\@secondoftwo\else\expandafter\@firstoftwo\fi} 
\makeatother

\newcommand{\steven}[1]{\textcolor{magenta}{\bf\small [#1 -- Steven]}}

\newcommand{\footnoteref}[1]{\textsuperscript{\ref{#1}}}

\newcommand*{\affmark}[1][*]{\textsuperscript{#1}}

\title{A Survey of Data Augmentation Approaches for NLP}

\author{Steven Y. Feng\bf{\thanks{\quad Equal contribution by the two authors.}}, \affmark[1] Varun Gangal\footnotemark[1], \affmark[1] Jason Wei\bf{\thanks{\quad AI Resident.}}, \affmark[2] Sarath Chandar,\affmark[3]\\{\bf Soroush Vosoughi},\affmark[4] {\bf Teruko Mitamura},\affmark[1] {\bf Eduard Hovy}\affmark[1]\\

  \affmark[1]Carnegie Mellon University, \affmark[2]Google Research\\ \affmark[3]Mila - Quebec AI Institute, \affmark[4]Dartmouth College
  \\
  {\tt \{syfeng,vgangal,teruko,hovy\}@cs.cmu.edu }\\
  \texttt{jasonwei@google.com} \hspace{3mm} \texttt{sarath.chandar@mila.quebec}\\
  \texttt{soroush@dartmouth.edu}\\  
}
\date{}

\usepackage{tikz-qtree}

\newcommand\ytl[2]{
\parbox[b]{8em}{\hfill{\color{cyan}\bfseries\sffamily #1}~$\cdots\cdots$~}\makebox[0pt][c]{$\bullet$}\vrule\quad \parbox[c]{14cm}{\vspace{7pt}\color{red!40!black!80}\raggedright\sffamily #2.\\[7pt]}\\[-3pt]}

\begin{document}
\maketitle
\setlength{\abovedisplayskip}{4pt}
\setlength{\belowdisplayskip}{4pt}
\begin{abstract}

Data augmentation has recently seen increased interest in NLP due to more work in low-resource domains, new tasks, and the popularity of large-scale neural networks that require large amounts of training data. 
Despite this recent upsurge, this area is still relatively underexplored, perhaps due to the challenges posed by the discrete nature of language data. In this paper, we present a comprehensive and unifying survey of data augmentation for NLP by summarizing the literature in a structured manner. We first introduce and motivate data augmentation for NLP, and then discuss major methodologically representative approaches. Next, we highlight techniques that are used for popular NLP applications and tasks. We conclude by outlining current challenges and directions for future research. Overall, our paper aims to clarify the landscape of existing literature in data augmentation for NLP and motivate additional work in this area. We also present a GitHub repository with a paper list that will be continuously updated at \url{https://github.com/styfeng/DataAug4NLP}.

\end{abstract}

\section{Introduction}
\label{sec:intro}
Data augmentation (\textsc{DA}) refers to strategies for increasing the diversity of training examples without explicitly collecting new data. It has received active attention in recent machine learning (ML) research in the form of well-received, general-purpose techniques such as \textsc{UDA} \cite{xie2019unsupervised} (\ref{subsec:rule-based}), which used \textit{backtranslation} \cite{sennrich-etal-2016-improving}, \textit{AutoAugment} \cite{cubuk2018autoaugment}, and \textit{RandAugment} \cite{cubuk2020randaugment}, and \textsc{MixUp} \cite{zhang2018mixup} (\ref{subsec:example_interpolation}). These are often first explored in computer vision (CV), and DA's adaptation for natural language processing (NLP) seems secondary and comparatively underexplored, perhaps due to challenges presented by the discrete nature of language, which rules out continuous noising and makes it more difficult to maintain invariance. 

Despite these challenges, there has been increased interest and demand for \textsc{DA} for NLP. As NLP grows due to off-the-shelf availability of large pretrained models, there are increasingly more tasks and domains to explore. Many of these are low-resource, and have a paucity of training examples, creating many use-cases for which \textsc{DA} can play an important role. Particularly, for many non-classification NLP tasks such as span-based tasks and generation, \textsc{DA} research is relatively sparse despite their ubiquity in real-world settings.

\begin{figure}
\begin{tabular}{@{}ll@{}}
\includegraphics[width=0.48\textwidth]{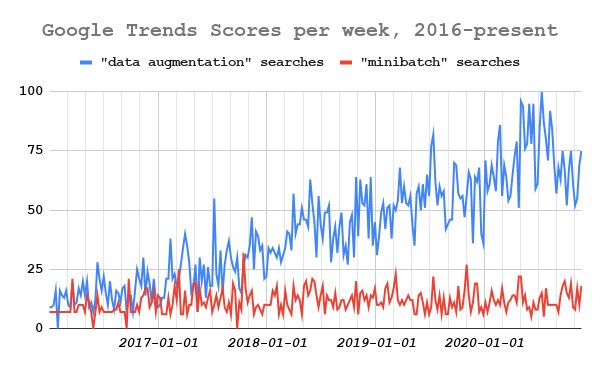}\\
\end{tabular}
\vspace{-0.8\abovedisplayskip}
  \caption{\label{tab:googleTrends} Weekly Google Trends scores for the search term \emph{"data augmentation"}, with a control, uneventful ML search term (\emph{"minibatch"}) for comparison.}
\vspace{-2ex}
\end{figure}

Our paper aims to sensitize the NLP community towards this growing area of work, 
which has also seen increasing interest in ML overall (as seen in Figure \ref{tab:googleTrends}). As interest and work on this topic continue to increase, this is an opportune time for a paper of our kind to (i) give a bird's eye view of DA for NLP, and (ii) identify key challenges to effectively motivate and orient interest in this area. To the best of our knowledge, this is the first survey to take a detailed look at \textsc{DA} methods for NLP.\footnote{\citet{9240734} present a smaller-scale text data augmentation survey that is concise and focused. Our work serves as a more comprehensive survey with larger coverage and is more up-to-date.}




This paper is structured as follows. Section \ref{sec:background} discusses what DA is, 
its goals and trade-offs, and why it works. Section \ref{sec:techniquesandmethods} describes popular methodologically representative DA techniques for NLP
---which we categorize into rule-based (\ref{subsec:rule-based}), example interpolation-based (\ref{subsec:example_interpolation}), or model-based (\ref{subsec:model_based}). Section \ref{sec:applications} discusses useful NLP applications for DA, including low-resource languages (\ref{applications:1}), mitigating bias (\ref{applications:2}), fixing class imbalance (\ref{applications:3}), few-shot learning (\ref{applications:4}), and adversarial examples (\ref{applications:5}). Section \ref{sec:tasks} describes DA methods for common NLP tasks including 
summarization (\ref{tasks:3}), question answering (\ref{tasks:5}), sequence tagging tasks (\ref{tasks:4}), parsing tasks (\ref{tasks:7}), grammatical error correction (\ref{tasks:9}), neural machine translation (\ref{tasks:10}), data-to-text NLG (\ref{tasks:8}), open-ended and conditional text generation (\ref{tasks:1}), dialogue (\ref{tasks:2}), and multimodal tasks (\ref{tasks:11}). Finally, Section \ref{sec:currentchallenges} discusses challenges and future directions in DA for NLP. Appendix \ref{sec:appendix_blog_posts} lists useful blog posts and code repositories.

Through this work, we hope to emulate past papers which have surveyed \textsc{DA} methods for other types of data, such as images \cite{shorten2019survey}, faces \cite{wang2019survey}, and time series \cite{iwana2020empirical}. We hope to draw further attention, elicit broader interest, and motivate additional work in \textsc{DA}, particularly for NLP.

\section{Background}
\label{sec:background}
\noindent \textbf{What is data augmentation?} \label{whatis}
Data augmentation (\textsc{DA}) encompasses methods of increasing training data diversity without directly collecting more data. Most strategies either add slightly modified copies of existing data or create synthetic data, aiming for the augmented data to act as a regularizer and reduce overfitting when training ML models \cite{shorten2019survey,garcia2020data}. DA has been commonly used in CV, where techniques like \textit{cropping}, \textit{flipping}, and \textit{color jittering} are a standard component of model training. 
In NLP, where the input space is discrete, how to generate effective augmented examples that capture the desired invariances is less obvious.

\noindent \paragraph{What are the goals and trade-offs?} \label{goals}
Despite challenges associated with text, many DA techniques for NLP have been proposed, ranging from rule-based manipulations \cite{10.5555/2969239.2969312} to more complicated generative approaches \cite{liu2020databoost}. As DA aims to provide an alternative to collecting more data, an ideal DA technique should be both easy-to-implement and improve model performance. 
Most offer trade-offs between these two. 

Rule-based techniques are easy-to-implement but usually offer incremental performance improvements \cite{li-etal-2017-robust,wei-zou-2019-eda,wei2021text}. Techniques leveraging trained models 
may be more costly to implement but introduce more data variation, leading to better performance boosts. Model-based techniques customized for downstream tasks can have strong  
effects on performance but be difficult to develop and utilize. 

Further, the distribution of augmented data should neither be too similar nor too different from the original. This may lead to greater overfitting or poor performance through training on examples not representative of the given domain, respectively. Effective DA approaches should aim for a balance.

\citet{kashefi-hwa-2020-quantifying} devise a KL-Divergence-based unsupervised procedure to \emph{preemptively} choose among \textsc{DA} heuristics, rather than a typical "run-all-heuristics" comparison, which can be very time and cost intensive.

\noindent \paragraph{Interpretation of DA} \label{why}
\citet{dao2019kernel} note that \textit{"data augmentation is typically performed in an ad-hoc manner with little understanding of the underlying theoretical principles"}, and claim the typical explanation of \textsc{DA} as \textit{regularization} to be insufficient. Overall, there indeed appears to be a lack of research on \textit{why} exactly DA works. Existing work on this topic is mainly surface-level, and rarely investigates the theoretical underpinnings and principles. We discuss this challenge more in \S \ref{sec:currentchallenges}, and highlight some of the existing work below.

\citet{bishop1995training} show training with noised examples is reducible to Tikhonov regularization (subsumes L2). \citet{rajput2019does} 
show that DA can increase the positive margin for classifiers, but only when augmenting exponentially many examples for common DA methods.

\citet{dao2019kernel} think of DA transformations as kernels, and find two ways DA helps: averaging of features and variance regularization. \citet{chen2020group} show that DA leads to variance reduction by averaging over orbits of the group that keep the data distribution approximately invariant. 

\section{Techniques \& Methods}
\label{sec:techniquesandmethods}
We now discuss some methodologically representative DA techniques which are relevant to all tasks via the extensibility of their formulation.\footnote{\label{note1}Table \ref{table:comparison} compares several \textsc{DA} methods by various aspects relating to their applicability, dependencies, and requirements.}

\subsection{Rule-Based Techniques}\label{subsec:rule-based}
Here, we cover \textsc{DA} primitives which use easy-to-compute, predetermined transforms sans model components. \textit{Feature space DA} approaches generate augmented examples in the model's feature space rather than input data. Many few-shot learning approaches \cite{hariharan2017low,schwartz2018delta} leverage estimated feature space "analogy" transformations between examples of known classes to augment for novel classes (see \S\ref{applications:4}). \citet{paschali2019data} use iterative affine transformations and projections to maximally "stretch" an example along the class-manifold.

\citet{wei-zou-2019-eda} propose \textsc{Easy Data Augmentation (EDA)}, a set of token-level random perturbation operations including 
\textit{random insertion, deletion}, and \textit{swap}.
They show improved performance on many text classification tasks. UDA \cite{xie2019unsupervised} show how supervised \textsc{DA} methods can be exploited for unsupervised data through consistency training on $(x,DA(x))$ pairs.

\begin{figure}
\centering
\begin{tabular}{@{}ll@{}}
\includegraphics[width=0.40\textwidth]{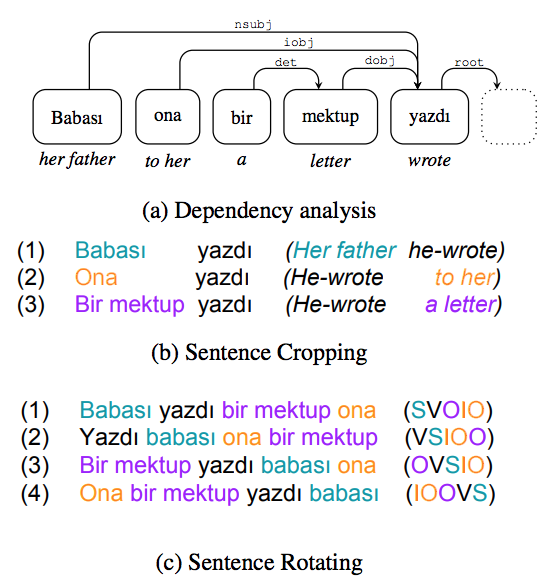}\\
\end{tabular}
\vspace{-0.8\abovedisplayskip}
  \caption{\label{img-deptreemorph} \emph{Dependency tree morphing} \textsc{DA} applied to a Turkish sentence, \citet{csahin2019data}}
\vspace{-2ex}
\end{figure}

For paraphrase identification, \citet{chen-etal-2020-finding} construct a signed graph over the data, with individual sentences as nodes and pair labels as signed edges. They use balance theory and transitivity to infer augmented sentence pairs from this graph. Motivated by image cropping and rotation, \citet{csahin2019data} propose \emph{dependency tree morphing}. For dependency-annotated sentences, children of the same parent are swapped (à la rotation) or some deleted (à la cropping), as seen in Figure \ref{img-deptreemorph}. This is most beneficial for language families with rich case marking systems (e.g. \textit{Baltic} and \textit{Slavic}).

\subsection{Example Interpolation Techniques}\label{subsec:example_interpolation}
Another class of \textsc{DA} techniques, pioneered by \textsc{MixUp} \cite{zhang2018mixup}, interpolates the inputs and labels of two or more real examples. 
This class of techniques is also sometimes referred to as \emph{Mixed Sample Data Augmentation} (MSDA). Ensuing work has explored interpolating inner components \cite{verma2019manifold,faramarzi2020patchup}, more general mixing schemes \cite{guo2020nonlinear}, and adding adversaries \cite{beckham2019adversarial}.

Another class of extensions of \textsc{MixUp} which has been growing in the vision community attempts to fuse raw input image pairs together into a single input image, rather than improve the continuous interpolation mechanism. Examples of this paradigm include \textsc{CutMix} \cite{yun2019cutmix}, \textsc{CutOut} \cite{devries2017improved} and \textsc{Copy-Paste} \cite{ghiasi2020simple}. For instance, \textsc{CutMix} replaces a small sub-region of Image A with a patch sampled from Image B, with the labels mixed in proportion to sub-region sizes. There is potential to borrow ideas and inspiration from these works for NLP, e.g. for multimodal work involving both images and text (see \textit{"Multimodal challenges"} in \S\ref{sec:currentchallenges}).

A bottleneck to using \textsc{MixUp} for NLP tasks was the requirement of continuous inputs. This has been overcome by 
mixing embeddings or higher hidden layers \cite{chen2020mixtext}. Later variants propose speech-tailored mixing schemes \cite{jindalspeechmix} and interpolation with adversarial examples \cite{cheng-etal-2020-advaug}, among others.

\textsc{Seq2MixUp} \cite{guo-etal-2020-sequence} generalizes \textsc{MixUp} for sequence transduction tasks in two ways - the "hard" version samples a binary mask (from a Bernoulli with a $\beta(\alpha,\alpha)$ prior) and picks from one of two sequences at each token position, while the "soft" version softly interpolates between sequences based on a coefficient sampled from $\beta(\alpha,\alpha)$. The "soft" version is found to outperform the "hard" version and earlier interpolation-based techniques like \textsc{SwitchOut} \cite{wang-etal-2018-switchout}.

\subsection{Model-Based Techniques}\label{subsec:model_based}
Seq2seq and language models have also been used for \textsc{DA}. The popular \textsc{backtranslation} method \cite{sennrich-etal-2016-improving} translates a sequence into another language and then back into the original language. \citet{kumar-etal-2019-submodular} train seq2seq models with their proposed method DiPS which learns to generate diverse paraphrases of input text using a modified decoder with a submodular objective, and show its effectiveness as DA for several classification tasks. Pretrained language models such as RNNs \cite{kobayashi-2018-contextual} and transformers \cite{yang2020g} have also been used for augmentation.

\begin{figure}
\centering
\begin{tabular}{@{}ll@{}}
\includegraphics[width=0.36\textwidth]{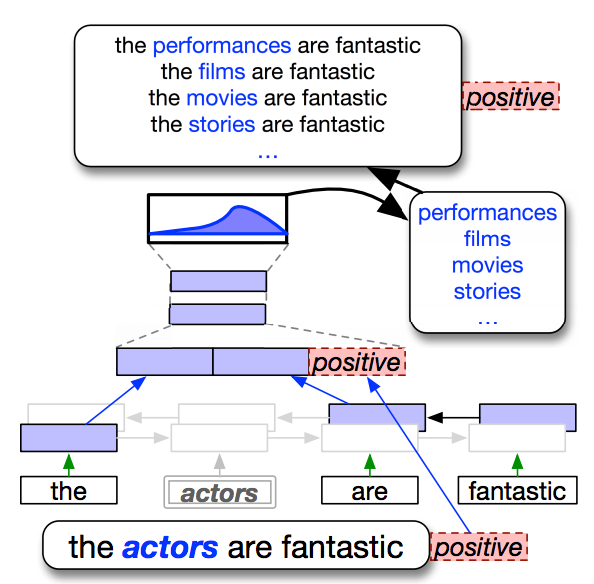}\\
\end{tabular}
\vspace{-0.8\abovedisplayskip}
  \caption{\label{tab:qualitative_1} \emph{Contextual Augmentation}, \citet{kobayashi-2018-contextual}}
\vspace{-2ex}
\end{figure}

\citet{kobayashi-2018-contextual} generate augmented examples by replacing words with others randomly drawn according to the recurrent language model's distribution based on the current context (illustration in Figure \ref{tab:qualitative_1}). \citet{yang2020g} propose \textsc{G-DAug}$^c$ which generates synthetic examples using pretrained transformer language models, and selects the most informative and diverse set for augmentation. \citet{gao-etal-2019-soft} advocate retaining the full distribution through "soft" augmented examples, showing gains on machine translation.

\citet{nie-etal-2020-named} augment word representations with a context-sensitive attention-based mixture of their semantic neighbors from a pretrained embedding space, and show its effectiveness for NER on social media text. Inspired by denoising autoencoders, \citet{ng-etal-2020-ssmba} use a corrupt-and-reconstruct approach, with the corruption function $q(x'|x)$ masking an arbitrary number of word positions and the reconstruction function $r(x|x')$ unmasking them using BERT \cite{devlin2019bert}. Their approach works well on domain-shifted 
test sets across 9 datasets on sentiment, NLI, and NMT.

\citet{feng-etal-2019-keep} propose a task called \textsc{Semantic Text Exchange} (STE) which involves adjusting the overall semantics of a text to fit the context of a new word/phrase that is inserted called the \textit{replacement entity (RE)}. They do so by using a system called SMERTI and a masked LM approach. 
While not proposed directly for DA, it can be used as such, as investigated in \citet{feng2020genaug}.

Rather than starting from an existing example and modifying it, some model-based \textsc{DA} approaches directly estimate a generative process from the training set and sample from it. \citet{anaby2020not} learn a label-conditioned generator by finetuning GPT-2 \cite{radford2019language} on the training data, using this to generate candidate examples per class. A classifier trained on the original training set is then used to select top $k$ candidate examples which confidently belong to the respective class for augmentation. \citet{quteineh2020textual} use a similar label-conditioned GPT-2 generation method, and demonstrate its effectiveness as a \textsc{DA} method in an active learning setup.

Other approaches include syntactic or controlled paraphrasing \cite{iyyer2018adversarial,kumar-etal-2020-syntax}, document-level paraphrasing \cite{gangal2021nareor}, augmenting misclassified examples \cite{dreossi2018counterexample}, BERT cross-encoder labeling of new inputs \cite{thakur2021augmented}, guided generation using large-scale generative language models \cite{liu2020databoost,liu2020offensive}, and \textit{automated text augmentation} \cite{bowen_automated,cai-etal-2020-data}. Models can also learn to combine together simpler DA primitives \cite{cubuk2018autoaugment,ratner2017learning} or add human-in-the-loop \cite{kaushik2020learning,kaushik2021explaining}.

\setlength{\tabcolsep}{2.4pt}
\begin{table*}[!ht]
    \centering
    \footnotesize
    \begin{tabular}{l|cccccc}
    \toprule
    \textbf{\textsc{DA} Method} & Ext.Know & Pretrained & Preprocess & Level  & Task-Agnostic    \\ 
    \midrule
    \textsc{Synonym Replacement} \cite{10.5555/2969239.2969312} & \ding{51} & $\times$ & tok & Input  & \ding{51}   \\
    \textsc{Random Deletion} \cite{wei-zou-2019-eda} & $\times$ & $\times$ & tok & Input  & \ding{51}   \\
    \textsc{Random Swap} \cite{wei-zou-2019-eda} & $\times$ & $\times$ & tok & Input  & \ding{51}   \\
        \textsc{Backtranslation} \cite{sennrich-etal-2016-improving} & $\times$ & \ding{51} & Depends & Input  & \ding{51}   \\
        \textsc{SCPN} \cite{wieting-gimpel-2017-revisiting} & $\times$ & \ding{51} & const & Input  & \ding{51}   \\
    \textsc{Semantic Text Exchange} \cite{feng-etal-2019-keep} & $\times$ & \ding{51} & const & Input  & \ding{51}   \\
    \textsc{ContextualAug} \cite{kobayashi-2018-contextual} & $\times$ & \ding{51} & - & Input  & \ding{51}   \\
    \textsc{LAMBADA} \cite{anaby2020not} & $\times$ & \ding{51} & - & Input & $\times$   \\
    \textsc{GECA} \cite{andreas-2020-good} & $\times$  & $\times$ & tok & Input  & $\times$ \\
       \textsc{SeqMixUp}  \cite{guo-etal-2020-sequence} & $\times$ & $\times$ & tok & Input & $\times$ \\
    \textsc{SwitchOut}  \cite{wang2018switchout} & $\times$  & $\times$ & tok & Input  & $\times$ \\   
    \textsc{Emix}  \cite{jindal-etal-2020-augmenting} & $\times$  & $\times$ & - & Emb/Hidden  & \ding{51} \\
    \textsc{SpeechMix}  \cite{jindalspeechmix} & $\times$  & $\times$ & - & Emb/Hidden  & Speech/Audio \\
        \textsc{MixText}  \cite{chen2020mixtext} & $\times$  & $\times$ & - & Emb/Hidden  & \ding{51} \\
   \textsc{SignedGraph}  \cite{chen-etal-2020-finding} & $\times$  & $\times$ & - & Input  & $\times$ \\
      \textsc{DTreeMorph}  \cite{csahin2019data} & $\times$  & $\times$ & dep & Input  & \ding{51} \\
            \textsc{$Sub^2$}  \cite{shi2101substructure} & $\times$  & $\times$ & dep & Input  & Substructural \\
\textsc{DAGA}  \cite{ding-etal-2020-daga} & $\times$  & $\times$ & tok & Input+Label  & $\times$ \\
\textsc{WN-Hypers}  \cite{feng2020genaug} & \ding{51}  & $\times$ & const+KWE & Input  & \ding{51} \\
\textsc{Synthetic Noise}  \cite{feng2020genaug} & $\times$  & $\times$ & tok & Input  & \ding{51} \\
\textsc{UEdin-MS} (DA part) \cite{grundkiewicz-etal-2019-neural} & \ding{51}  & $\times$ & tok & Input & \ding{51} \\
\textsc{Nonce} \cite{gulordava-etal-2018-colorless} & \ding{51} & $\times$ & const & Input & \ding{51} \\
\textsc{XLDA} \cite{singh2019xlda} & $\times$ & \ding{51} & Depends & Input & \ding{51} \\
\textsc{SeqMix} \cite{zhang-etal-2020-seqmix} & $\times$ & \ding{51} & tok & Input+Label & $\times$ \\
\textsc{Slot-Sub-LM} \cite{louvan2020simple} & $\times$ & \ding{51} & tok & Input & \ding{51} \\
\textsc{UBT \& TBT} \cite{vaibhav-etal-2019-improving} & $\times$ & \ding{51} & Depends & Input & \ding{51} \\
\textsc{Soft Contextual DA} \cite{gao-etal-2019-soft} & $\times$ & \ding{51} & tok & Emb/Hidden & \ding{51} \\
\textsc{Data Diversification} \cite{nguyen2020data} & $\times$ & \ding{51} & Depends & Input & \ding{51} \\
\textsc{DiPS} \cite{kumar-etal-2019-submodular} & $\times$ & \ding{51} & tok & Input & \ding{51} \\
\textsc{Augmented SBERT} \cite{thakur2021augmented} & $\times$ & \ding{51} & - & Input+Label & Sentence Pairs \\
    \bottomrule
    \end{tabular}
    \caption{Comparing a selection of \textsc{DA} methods by various aspects relating to their applicability, dependencies, and requirements. \emph{Ext.Know}, \emph{KWE}, \emph{tok}, \emph{const}, and \emph{dep} stand for External Knowledge, keyword extraction, tokenization, constituency parsing, and dependency parsing, respectively. \emph{Ext.Know} refers to whether the DA method requires external knowledge (e.g. WordNet) and \emph{Pretrained} if it requires a pretrained model (e.g. BERT). \emph{Preprocess} denotes preprocessing required, \emph{Level} denotes the depth at which data is modified by the DA, and \emph{Task-Agnostic} refers to whether the DA method can be applied to different tasks. See Appendix \ref{sec:appendix_table_typology} for further explanation.}
    \textbf{\label{table:comparison}}
    \vspace{-5ex}
\end{table*}

\section{Applications}
\label{sec:applications}
In this section, we discuss several DA methods for some common NLP applications.\footnoteref{note1}

\subsection{Low-Resource Languages}\label{applications:1} Low-resource languages are an important and challenging application for DA, typically for neural machine translation (NMT). Techniques using external knowledge such as WordNet \cite{miller1995wordnet} may be difficult to use effectively here.\footnote{Low-resource language challenges discussed more in \S\ref{sec:currentchallenges}.} There are ways to leverage high-resource languages for low-resource languages, particularly if they have similar linguistic properties. \citet{xia-etal-2019-generalized} use this approach to improve low-resource NMT.

\citet{info11050255} use \textit{backtranslation} and self-learning to generate augmented training data. Inspired by work in CV, \citet{fadaee-etal-2017-data} generate additional training examples that contain low-frequency (rare) words in synthetically created contexts. \citet{ijcai2020-533} present a DA framework to generate multi-lingual code-switching data to finetune multilingual-BERT. It encourages the alignment of representations from source and multiple target languages once by mixing their context information. They see improved performance across 5 tasks with 19 languages.

\subsection{Mitigating Bias}\label{applications:2} 
\citet{zhao-etal-2018-gender} attempt to mitigate gender bias in coreference resolution by creating an augmented dataset identical to the original but biased towards the underrepresented gender (using gender swapping of entities such as replacing \textit{"he"} with \textit{"she"}) and train on the union of the two datasets. \citet{lu2019gender} formally propose \textsc{counterfactual DA} (CDA) for gender bias mitigation, which involves causal interventions that break associations between gendered and gender-neutral words. \citet{zmigrod-etal-2019-counterfactual} and \citet{hall-maudslay-etal-2019-name} propose further improvements to \textsc{CDA}. \citet{moosavi2020improving} augment training sentences with their corresponding predicate-argument structures, improving the robustness of transformer models against various types of biases.

\subsection{Fixing Class Imbalance}\label{applications:3} Fixing class imbalance typically involves a combination of undersampling and oversampling. 
\textsc{Synthetic Minority Oversampling Technique} (SMOTE) \cite{chawla2002smote}, which generates augmented minority class examples through interpolation, still remains popular \cite{fernandez2018smote}. \textsc{Multilabel SMOTE} (MLSMOTE) \cite{MLSmote} modifies SMOTE to balance classes for multi-label classification, where classifiers predict more than one class at the same time. Other techniques such as EDA \cite{wei-zou-2019-eda} can possibly be used for oversampling as well. 

\subsection{Few-Shot Learning}\label{applications:4} DA methods can ease few-shot learning by adding more examples for novel classes introduced in the few-shot phase. \citet{hariharan2017low} use learned analogy transformations $\phi(z_1,z_2,x)$ between example pairs from a non-novel class $z_{1} \rightarrow z_{2}$ to generate augmented examples $x \rightarrow x'$ for novel classes. 
\citet{schwartz2018delta} generalize this to beyond just linear offsets, through their "$\Delta$-network" autoencoder which learns the distribution $P(z_2|z_1,C)$ from all $y^{*}_{z_1} = y^{*}_{z_2} = C$ pairs, where $C$ is a class and $y$ is the ground-truth labelling function. Both these methods are applied only on image tasks, but their theoretical formulations are generally applicable, and hence we discuss them.

\citet{kumar2019closer} apply these and other DA methods for few-shot learning of novel intent classes in task-oriented dialog.
\citet{wei2021few} show that data augmentation facilitates curriculum learning for training triplet networks for few-shot text classification. \citet{lee2021neural} use T5 to generate additional examples for data-scarce classes.

\subsection{Adversarial Examples (AVEs)}\label{applications:5} Adversarial examples can be generated using innocuous label-preserving transformations (e.g. paraphrasing) that fool state-of-the-art NLP models, as shown in \citet{jia2019certified}. Specifically, they add sentences with distractor spans to passages to construct AVEs for span-based QA. \citet{zhang-etal-2019-paws} construct AVEs for paraphrase detection using word swapping. \citet{kang-etal-2018-adventure} and \citet{glockner-etal-2018-breaking} create AVEs for textual entailment using WordNet relations.

\section{Tasks}
\label{sec:tasks}
In this section, we discuss several DA works for common NLP tasks.\footnoteref{note1} We focus on non-classification tasks as classification is worked on by default, and well covered in earlier sections (e.g. \S\ref{sec:techniquesandmethods} and \S\ref{sec:applications}). Numerous previously mentioned DA techniques, e.g. \cite{wei-zou-2019-eda,chen-etal-2020-finding,anaby2020not}, have been used or can be used for text classification tasks.

\subsection{Summarization}\label{tasks:3}
\citet{fabbri2020improving} investigate \textit{backtranslation} as a DA method for few-shot abstractive summarization with the use of a consistency loss inspired by UDA. \citet{parida-motlicek-2019-abstract} propose an iterative DA approach for abstractive summarization that uses a mix of synthetic and real data, where the former is generated from Common Crawl. \citet{zhu2019transforming} introduce a query-focused summarization \cite{Dang2005OverviewOD} dataset collected using Wikipedia called \textsc{WikiRef} which can be used for DA. \citet{pasunuru2021data} use DA methods to construct two training datasets for Query-focused Multi-Document Summarization (QMDS) called \textsc{QmdsCnn} and \textsc{QmdsIr} by modifying CNN/DM \cite{48562} and mining search-query logs, respectively.

\subsection{Question Answering (QA)}\label{tasks:5}
\citet{longpre-etal-2019-exploration} investigate various DA and sampling techniques for domain-agnostic QA including paraphrasing by \textit{backtranslation}. \citet{yang2019data} propose a DA method using distant supervision to improve BERT finetuning for open-domain QA. \citet{riabi2020synthetic} leverage Question Generation models to produce augmented examples for zero-shot cross-lingual QA. \citet{singh2019xlda} propose \textsc{XLDA}, or \textsc{cross-lingual DA}, which substitutes a portion of the input text with its translation in another language, improving performance across multiple languages on NLI tasks including the SQuAD QA task. \citet{asai-hajishirzi-2020-logic} use logical and linguistic knowledge to generate additional training data to improve the accuracy and consistency of QA responses by models. \citet{wei2018fast} introduce a new QA architecture called QANet that shows improved performance on SQuAD when combined with augmented data generated using backtranslation.


\subsection{Sequence Tagging Tasks}\label{tasks:4}
\citet{ding-etal-2020-daga} propose \textsc{DAGA}, a two-step \textsc{DA} process. First, a language model over sequences of tags and words linearized as per a certain scheme 
is learned. Second, sequences are sampled from this language model and de-linearized to generate new examples. \citet{csahin2019data}, discussed in \S\ref{subsec:rule-based}, use \emph{dependency tree morphing} (Figure \ref{img-deptreemorph}) to generate additional training examples on the downstream task of part-of-speech (POS) tagging. 

\citet{dai-adel-2020-analysis} modify DA techniques proposed for sentence-level tasks for named entity recognition (NER), including label-wise token and synonym replacement, and show improved performance using both recurrent and transformer models. \citet{zhang-etal-2020-seqmix} propose a DA method based on \textsc{MixUp} called \textsc{SeqMix} for active sequence labeling by augmenting queried samples, showing improvements on NER and Event Detection. 

\subsection{Parsing Tasks}\label{tasks:7}
\citet{jia-liang-2016-data} propose \textsc{data recombination} for injecting task-specific priors to neural semantic parsers. A 
synchronous context-free grammar (SCFG) is induced from training data, and new "recombinant" examples are sampled. \citet{yu2020grappa} introduce \textsc{Grappa}, a pretraining approach for table semantic parsing, and generate synthetic question-SQL pairs via an SCFG. 
\citet{andreas-2020-good} use 
\textit{compositionality} to construct synthetic examples for downstream tasks like semantic parsing. Fragments of original examples are replaced with fragments from other examples in similar contexts. 

\citet{vania-etal-2019-systematic} investigate DA for low-resource dependency parsing including \emph{dependency tree morphing} from \citet{csahin2019data} (Figure \ref{img-deptreemorph}) and modified \textit{nonce} sentence generation from \citet{gulordava-etal-2018-colorless}, which replaces content words with other words of the same POS, morphological features, and dependency labels.


\subsection{Grammatical Error Correction (GEC)}\label{tasks:9}
Lack of parallel data is typically a barrier for GEC. Various works have thus looked at DA methods for GEC. 
We discuss some here, and more can be found in Table \ref{table:additional_DA_works_GEC} in Appendix \ref{section:appendix_additional_DA_works}. 

There is work that makes use of additional resources. \citet{boyd-2018-using} use German edits from Wikipedia revision history and use those relating to GEC as augmented training data.  
\citet{zhang2019sequencetosequence} explore multi-task transfer, or the use of annotated data from other tasks.

There is also work that adds synthetic errors to noise the text. 
\citet{wang2020controllable} investigate two approaches: token-level perturbations 
and training error generation models with a filtering strategy to keep generations with sufficient errors. \citet{grundkiewicz-etal-2019-neural} use \textit{confusion sets} generated by a spellchecker for noising. \citet{choe-etal-2019-neural} learn error patterns from small annotated samples along with \textit{POS-specific noising}.

There have also been approaches to improve the diversity of generated errors. \citet{wan-etal-2020-improving} investigate noising through editing the latent representations of grammatical sentences, and \citet{xie-etal-2018-noising} use a neural sequence transduction model and beam search noising procedures.

\subsection{Neural Machine Translation (NMT)}\label{tasks:10}
There are many works which have investigated DA for NMT. We highlighted some in \S\ref{sec:techniquesandmethods} and \S\ref{applications:1}, e.g. \cite{sennrich-etal-2016-improving,fadaee-etal-2017-data,xia-etal-2019-generalized}. 
We discuss some further ones here, and more can be found in Table \ref{table:additional_DA_works_MT} in Appendix \ref{section:appendix_additional_DA_works}.

\citet{wang-etal-2018-switchout} propose \textsc{SwitchOut}, a DA method that randomly replaces words in both source and target sentences with other random words from their corresponding vocabularies. \citet{gao-etal-2019-soft} introduce \textsc{Soft Contextual DA} that softly augments randomly chosen words in a sentence using a contextual mixture of multiple related words over the vocabulary. \citet{nguyen2020data} propose \textsc{Data Diversification} which merges original training data with the predictions of several forward and backward models.

\subsection{Data-to-Text NLG}\label{tasks:8}
\textit{Data-to-text NLG} refers to tasks which require generating natural language descriptions of structured or semi-structured data inputs, e.g. game score tables \cite{wiseman2017challenges}. Randomly perturbing game score values without invalidating overall game outcome is one \textsc{DA} strategy explored in game summary generation \cite{hayashi2019findings}.


Two popular recent benchmarks are E2E-NLG \cite{duvsek2018findings} and WebNLG \cite{gardent2017webnlg}. Both involve generation from structured inputs - meaning representation (MR) sequences and triple sequences, respectively. \citet{montella2020denoising} show performance gains on WebNLG by \textsc{DA} using Wikipedia sentences as targets and parsed OpenIE triples as inputs. \citet{tandon2018tnt} propose \textsc{DA} for E2E-NLG based on permuting the input MR sequence. \citet{kedzie2019good} inject Gaussian noise into a trained decoder's hidden states and sample diverse augmented examples from it. This 
sample-augment-retrain loop helps performance on E2E-NLG.  

\subsection{Open-Ended \& Conditional Generation}\label{tasks:1}
There has been limited work on DA for open-ended and conditional text generation. \citet{feng2020genaug} experiment with a suite of DA methods for finetuning GPT-2 on a low-resource domain in attempts to improve the quality 
of generated continuations, which they call \textsc{GenAug}. They find that \textsc{WN-Hypers} (WordNet hypernym replacement of keywords) and \textsc{Synthetic Noise} (randomly perturbing non-terminal characters in words) are useful, and the quality of generated text improves to a peak at $\approx$ 3x the original amount of training data.


\subsection{Dialogue}\label{tasks:2}
Most DA approaches for dialogue focus on task-oriented dialogue. We outline some below, and more can be found in Table \ref{table:additional_DA_works_dialogue} in Appendix \ref{section:appendix_additional_DA_works}.

\citet{Quan2019EffectiveDA} present sentence and word-level DA approaches for end-to-end task-oriented dialogue. \citet{louvan2020simple} propose \textsc{lightweight augmentation}, a set of word-span and sentence-level DA methods for low-resource slot filling and intent classification. 

\citet{hou-etal-2018-sequence} present a seq2seq DA framework to augment dialogue utterances for dialogue language understanding \cite{young-pomdp}, 
including a \textit{diversity rank} to produce diverse utterances. 
\citet{zhang2019taskoriented} propose 
\textsc{MADA} to generate diverse responses using the property that several valid responses exist for a dialogue context. 

There is also DA work for spoken dialogue. \citet{hou-etal-2018-sequence}, \citet{kim-etal-2019-data}, \citet{zhao2019data}, and \citet{yoo2018data} investigate DA methods for dialogue and \emph{spoken language understanding} (SLU), including generative latent variable models.

\subsection{Multimodal Tasks}\label{tasks:11}
DA techniques have also been proposed for multimodal tasks where aligned data for multiple modalities is required. We look at ones that involve language or text. Some are discussed below, and more can be found in Table \ref{table:additional_DA_works_multimodal} in Appendix \ref{section:appendix_additional_DA_works}.

Beginning with speech, \citet{wang2020data} propose a DA method to improve the robustness of downstream dialogue models to speech recognition errors. \citet{unknown} and \citet{renduchintala2018multimodal} propose DA methods for end-to-end \emph{automatic speech recognition} (ASR). 

Looking at images or video, \citet{9206007} learn a cross-modality matching network to produce synthetic image-text pairs for multimodal classifiers. 
\citet{atlihabert} explore DA methods such as synonym replacement and contextualized word embeddings augmentation using BERT for \emph{image captioning}. \citet{kafle-etal-2017-data}, \citet{yokota-nakayama-2018-augmenting}, and \citet{tang2020semantic} propose methods for \emph{visual QA} including question generation and adversarial examples. 

\section{Challenges \& Future Directions}
\label{sec:currentchallenges}

Looking forward, data augmentation faces substantial challenges, specifically for NLP, and with these challenges, new opportunities for future work arise.



\vspace{-0.5mm}
\paragraph{Dissonance between empirical novelties and theoretical narrative:}
There appears to be a conspicuous lack of research on \textit{why} DA works. Most studies might show empirically that a DA technique works and provide some intuition, but it is currently challenging to measure the goodness of a technique without resorting to a full-scale experiment. 
A recent work in vision \cite{gontijo2020affinity} has proposed that affinity (the distributional shift caused by DA) and diversity (the complexity of the augmentation) can predict DA performance, but it is unclear how these results might translate to NLP.

\vspace{-0.5mm}
\paragraph{Minimal benefit for pretrained models on in-domain data:}With the popularization of large pretrained language models, it has come to light that a couple previously effective \textsc{DA} techniques for certain English text classification tasks \cite{wei-zou-2019-eda,sennrich-etal-2016-improving} provide little benefit for models like BERT and RoBERTa, which already achieve high performance on in-domain text classification \cite{longpre2020effective}. One hypothesis is that using simple DA techniques provides little benefit when finetuning large pretrained transformers on tasks for which examples are well-represented in the pretraining data, but DA methods could still be effective when finetuning on tasks for which examples are scarce or out-of-domain compared with the training data. Further work could study under which scenarios data augmentation for large pretrained models is likely to be effective.



\paragraph{Multimodal challenges:} While there has been increased work in multimodal DA, as discussed in \S\ref{tasks:11}, effective DA methods for multiple modalities has been challenging. Many works focus on augmenting a single modality or multiple ones separately. 
For example, there is potential to further explore simultaneous image and text augmentation for image captioning, such as a combination of \textsc{CutMix} \cite{yun2019cutmix} and caption editing.

\paragraph{Span-based tasks} offer unique DA challenges as there are typically many correlated classification decisions. For example, random token replacement may be a locally acceptable DA method but possibly disrupt coreference chains for latter sentences. DA techniques here 
must take into account dependencies between different locations in the text.

\paragraph{Working in specialized domains} such as those with domain-specific vocabulary and jargon (e.g. \emph{medicine}) can present challenges. Many pretrained models and external knowledge (e.g. WordNet) cannot be effectively used. Studies have shown that DA becomes less beneficial when applied to out-of-domain data, likely because the distribution of augmented data can substantially differ from the original data \cite{zhang-etal-2019-editing, herzig-etal-2020-tapas, campagna2020zeroshot, zhong-etal-2020-grounded}.

\paragraph{Working with low-resource languages} may present similar difficulties as specialized domains. Further, DA techniques successful in the high-resource scenario may not be effective for low-resource languages that are of a different language family or very distinctive in linguistic and typological terms. For example, those which are language isolates or lack high-resource cognates. 


\paragraph{More vision-inspired techniques:} Although many NLP DA methods have been inspired by analogous approaches in CV, there is potential for drawing further connections. Many CV DA techniques motivated by real-world invariances (e.g. many angles of looking at the same object) may have similar NLP interpretations. For instance, \emph{grayscaling} could translate to 
toning down aspects of the text (e.g. plural to singular, \textit{"awesome"} → \textit{"good"}). Morphing a dependency tree 
could be analogous to rotating an image, and 
paraphrasing techniques may be analogous to changing perspective. For example, negative data augmentation (NDA) \cite{sinha2021negative} involves creating out-of-distribution samples. It has so far been exclusively explored for CV, but could be investigated for text.

\paragraph{Self-supervised learning:}
More recently, \textsc{DA} has been increasingly used as a key component of self-supervised learning, particularly in vision \cite{chen2020simple}. In NLP, BART \cite{lewis-etal-2020-bart} showed that predicting deleted tokens as a pretraining task can achieve similar performance as the masked LM, and \textsc{Electra} \cite{clark2020electra} found that pretraining by predicting corrupted tokens outperforms BERT given the same model size, data, and compute. We expect future work will continue exploring how to effectively manipulate text for both pretraining and downstream tasks.

\paragraph{Offline versus online data augmentation:} In CV, standard techniques such as cropping and rotations are typically done stochastically, allowing for DA to be incorporated elegantly into the training pipeline. 
In NLP, however, it is unclear how to include a lightweight code module to apply DA stochastically. 
This is because DA techniques for NLP often leverage external resources (e.g. a dictionary for token substitution or translation model for backtranslation) that are not easily transferable across training pipelines. 
Thus, a common practice for DA in NLP is to generate augmented data offline and store it as additional data to be loaded during training.\footnote{See Appendix \ref{sec:pedro}.} 
Future work on a lightweight module for online DA in NLP could be fruitful, though another challenge will be determining when such a module will be helpful, which---compared with CV, where invariances being imposed are well-accepted---can vary substantially across NLP tasks.

\paragraph{Lack of unification} is a challenge for the current literature on data augmentation for NLP, and popular methods are often presented in an auxiliary fashion. Whereas there are well-accepted frameworks for DA for CV (e.g. default augmentation libraries in PyTorch, \emph{RandAugment} \cite{cubuk2020randaugment}), there are no such "generalized" DA techniques for NLP. Further, we believe that DA research would benefit from the establishment of standard and unified benchmark tasks and datasets to compare different augmentation methods.

\paragraph{Good data augmentation practices} would help make DA work more accessible and reproducible to the NLP and ML communities. On top of unified benchmark tasks, datasets, and frameworks/libraries mentioned above, other good practices include making code and augmented datasets publicly available, reporting variation among results (e.g. standard deviation across random seeds), and more standardized evaluation procedures. Further, transparent hyperparameter analysis, explicitly stating failure cases of proposed techniques, and discussion of the intuition and theory behind them would further improve the transparency and interpretability of DA techniques.



\section{Conclusion}
\label{sec:conclusions}
In this paper, we presented a comprehensive and structured survey of data augmentation for natural language processing (NLP). We provided a background about data augmentation and how it works, discussed major methodologically representative data augmentation techniques for NLP, and touched upon data augmentation techniques for popular NLP applications and tasks. Finally, we outlined current challenges and directions for future research, and showed that there is much room for further exploration. Overall, we hope our paper can serve as a guide for NLP researchers to decide on which data augmentation techniques to use, and inspire additional interest and work in this area. Please see the corresponding GitHub repository at \url{https://github.com/styfeng/DataAug4NLP}.








\bibliographystyle{acl_natbib}
\bibliography{anthology,acl2021}

\bigbreak
\appendix
\section*{Appendices}
\section{Useful Blog Posts and Code Repositories}\label{sec:appendix_blog_posts}
The following blog posts and code repositories could be helpful in addition to the information presented and papers/works mentioned in the body:
\begin{itemize}[leftmargin=*,nolistsep]
    \item Introduction to popular text augmentation techniques: \url{https://towardsdatascience.com/data-augmentation-in-nlp-2801a34dfc28}
    \item Detailed blog post on various text DA techniques: \url{https://amitness.com/2020/05/data-augmentation-for-nlp/}
    \item Lightweight library for DA on text and audio: \url{https://github.com/makcedward/nlpaug}
    \item \texttt{python} framework for adversarial examples: \url{https://github.com/QData/TextAttack}
\end{itemize}

\section{\textsc{DA} Methods Table - Description of Columns and Attributes}\label{sec:appendix_table_typology}
Table \ref{table:comparison} in the main body compares a non-exhaustive selection of \textsc{DA} methods along various aspects relating to their applicability, dependencies, and requirements. Below, we provide a more extensive description of each of this table's columns and their attributes.

\begin{enumerate}
    \item \textbf{Ext.Know}: Short for external knowledge, this column is \ding{51} when the data augmentation process requires knowledge resources which go beyond the immediate input examples and the task definition, such as WordNet \cite{miller1995wordnet} or PPDB \cite{pavlick-etal-2015-ppdb}. Note that we exclude the case where these resources are pretrained models under a separate point (next) for clarity, since these are widespread enough to merit a separate category.
    \item \textbf{Pretrained}: Denotes that the data augmentation process requires a pretrained model, such as BERT \cite{devlin2019bert} or GPT-2 \cite{radford2019language}. 
    \item \textbf{Preprocess}: Denotes the preprocessing steps, e.g. tokenization (\textit{tok}), dependency parsing (\textit{dep}), etc. required for the DA process. A hyphen (-) means either no preprocessing is required or that it was not explicitly stated.
    \item \textbf{Level}: Denotes the depth and extent to which elements of the instance/data are modified by the \textsc{DA}. Some primitives modify just the \textsc{Input} (e.g. word swapping), some modify both \textsc{Input} and \textsc{Label} (e.g. negation), while others make changes in the embedding or hidden space (\textsc{Embed/Hidden}) or higher representation layers enroute to the task model.
    \item \textbf{Task-Agnostic}: This is an approximate, partially subjective column denoting the extent to which a \textsc{DA} method can be applied to different tasks. When we say \ding{51} here, we don't denote a very rigid sense of the term task-agnostic, but mean that it would possibly easily extend to most NLP tasks as understood by the authors. Similarly, an $\times$ denotes being restricted to a specific task (or small group of related tasks) only. There can be other labels, denoting applicability to broad task families. For example, \textsc{substructural} denotes the family of tasks where sub-parts of the input are also valid input examples in their own right, e.g. constituency parsing. \textsc{Sentence Pairs} denotes tasks which involve pairwise sentence scoring such as paraphrase identification, duplicate question detection, and semantic textual similarity.
\end{enumerate}

\section{Additional DA Works by Task}\label{section:appendix_additional_DA_works}
See Table \ref{table:additional_DA_works_GEC} for additional DA works for GEC, Table \ref{table:additional_DA_works_MT} for additional DA works for neural machine translation, Table \ref{table:additional_DA_works_dialogue} for additional DA works for dialogue, and Table \ref{table:additional_DA_works_multimodal} for additional DA works for multimodal tasks. Each work is described briefly.

\section{Additional Figure}\label{sec:pedro}

\begin{figure}[ht]
\includegraphics[width=0.48\textwidth]{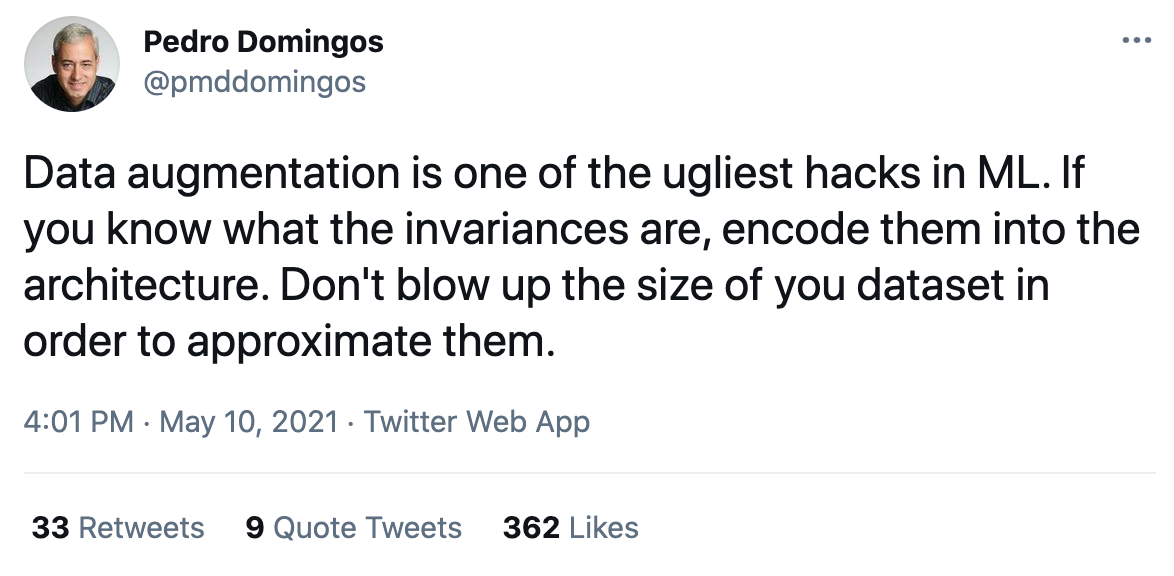}\\
\vspace{-0.8\abovedisplayskip}
  \caption{\label{tab:pedro} Pedro Domingos' quip about offline data augmentation.}
\end{figure}

\setlength{\tabcolsep}{1pt}
\begin{table*}[!ht]
    \centering
    \footnotesize
    \begin{tabular}{p{0.25\linewidth}|p{0.74\linewidth}}
    \toprule
    \textbf{Paper/Work} & \textbf{Brief Description} \\ 
    \midrule
    \citet{lichtarge-etal-2019-corpora} & Generate synthetic noised examples of Wikipedia sentences using backtranslation through various languages. \\
    \hline
    \citet{white-rozovskaya-2020-comparative} & Detailed comparative study of the DA for GEC systems UEdin-MS \cite{grundkiewicz-etal-2019-neural} and Kakao\&Brain \cite{choe-etal-2019-neural}. \\
    \hline
    \citet{foster-andersen-2009-generrate} & Introduces error generation tool called GenERRate which learns to generate ungrammatical text with various errors by using an error analysis file. \\
    \hline
    \citet{kimn-Japanese-GEC} & Use a set of syntactic rules for common Japanese grammatical errors to generate augmented error-correct sentence pairs for Japanese GEC.\\
    \hline
    \citet{Felice2016ArtificialEG} & Thesis that surveys previous work on error generation and investigates some new approaches using random and probabilistic methods.\\
    \hline
    \citet{xu-etal-2019-erroneous} & Noises using five error types: concatenation, misspelling, substitution, deletion, and transposition. Decent performance on the BEA 2019 Shared Task.\\
    \hline
    \citet{zhang2019sequencetosequence} & Explore backtranslation and feature discrimination for DA.\\
    \hline
    \citet{mizumoto-etal-2011-mining} & DA by extracting Japanese GEC training data from the revision log of a language learning SNS.\\
    \bottomrule
    \end{tabular}
    \caption{Additional DA works for grammatical error correction (GEC), along with a brief description of each.}
    \textbf{\label{table:additional_DA_works_GEC}}
    \vspace{-5ex}
\vspace{2cm}
    \centering
    \footnotesize
    \begin{tabular}{p{0.25\linewidth}|p{0.74\linewidth}}
    \toprule
    \textbf{Paper/Work} & \textbf{Brief Description} \\ 
    \midrule
    \citet{vaibhav-etal-2019-improving} & Present a \textit{synthetic noise induction} model which heuristically adds social media noise to text, and \textit{labeled backtranslation}.\\
    \hline
    \citet{hassan2017synthetic} & Present a DA method to project words from closely-related high-resource languages to low-resource languages using word embedding representations.\\
    \hline
    \citet{cheng-etal-2020-advaug} & Propose \textit{AdvAug}, an adversarial augmentation method for NMT, by sampling adversarial examples from a new vicinity distribution and using their embeddings to augment training.\\
    \hline
    \citet{graca-etal-2019-generalizing} & Investigate improvements to sampling-based approaches and the synthetic data generated by backtranslation.\\
    \hline
    \citet{bulte-tezcan-2019-neural} & Propose DA approaches for NMT that leverage information retrieved from a Translation Memory (TM) and using fuzzy TM matches.\\
    \hline
    \citet{KGNMT} & Propose an NMT model \textsc{KG-NMT} which is augmented by knowledge graphs to enhance semantic feature extraction and hence the translation of entities and terminological expressions.\\
    \hline
    \citet{peng2020dictionarybased} & Propose dictionary-based DA (\textsc{DDA}) for cross-domain NMT by synthesizing a domain-specific dictionary and automatically generating a pseudo in-domain parallel corpus.\\
    \hline
    \citet{li2020sentence} & Present a DA method using sentence boundary segmentation to improve the robustness of NMT on ASR transcripts.\\
    \hline
    \citet{multi-source-NMT} & Introduce DA methods for multi-source NMT that fills in incomplete portions of multi-source training data.\\
    \hline
    \citet{sugiyama-yoshinaga-2019-data} & Investigate effectiveness of DA by backtranslation for context-aware NMT.\\
    \hline
    \citet{li-specia-2019-improving} & Present DA methods to improve NMT robustness to noise while keeping models small, and explore the use of noise from external data (speech transcripts).\\
    \hline
    \citet{chinea-rios-etal-2017-adapting} & Propose DA method to create synthetic data by leveraging the embedding representation of sentences.\\
    \hline
    \citet{alves:hal-02907053} & Propose two methods for pipeline-based speech translation through the introduction of errors through 1. utilizing a speech processing workflow and 2. a rule-based method.\\
    \hline
    \citet{Kang2019VaLaRN} & Investigate extremely low-resource settings for NMT and a DA approach using a noisy dictionary and language models.\\
    \hline
    \citet{ijcai2020-496} & Investigate a DA method for lexically constraint-aware NMT to construct constraint-aware synthetic training data.\\
    \hline
    \citet{info11050255} & Propose a diversity DA method for low-resource NMT by generating diverse synthetic parallel data on both source and target sides using a restricted sampling strategy during decoding.\\
    \hline
    \citet{duan2020syntaxaware} & Propose syntax-aware DA methods with sentence-specific word selection probabilities using dependency parsing.\\
    \bottomrule
    \end{tabular}
    \caption{Additional DA works for neural machine translation (NMT), along with a brief description of each.}
    \textbf{\label{table:additional_DA_works_MT}}
    \vspace{-5ex}
\end{table*}

\clearpage
\setlength{\tabcolsep}{1pt}
\begin{table*}[ht!]
    \centering
    \footnotesize
    \begin{tabular}{p{0.15\linewidth}|p{0.84\linewidth}}
    \toprule
    \textbf{Paper/Work} & \textbf{Brief Description} \\ 
    \midrule
    \citet{gao2020paraphrase} & Propose a paraphrase augmented response generation (PARG) framework to improve dialogue generation by automatically constructing augmented paraphrased training examples based on dialogue state and act labels. \\
    \hline
    \citet{gritta2020conversation} & Introduce a graph-based representation of dialogues called Conversation Graph (ConvGraph) that can be used for DA by creating new dialogue paths. \\
    \hline
    \citet{yin2019dialog} & Propose an RL-based DA approach for dialogue state tracking (DST). \\
    \hline
    \citet{song2020data} & Propose a simple DA algorithm to improve the training of copy-mechanism models for dialogue state tracking (DST). \\
    \bottomrule
    \end{tabular}
    \caption{Additional DA works for dialogue, along with a brief description of each.}
    \textbf{\label{table:additional_DA_works_dialogue}}
    \vspace{-5ex}
    \vspace{2cm}
    \centering
    \footnotesize
    \begin{tabular}{p{0.15\linewidth}|p{0.84\linewidth}}
    \toprule
    \textbf{Paper/Work} & \textbf{Brief Description} \\ 
    \midrule
    \citet{10.1145/3266302.3266304} & Propose a DA method for emotion recognition from a combination of audio, visual, and textual modalities.\\
    \hline
    \citet{mou2020multimodal} & Introduce a DA method for Audio-Video Scene-Aware Dialogue, which involves dialogue containing a sequence of QA pairs about a video.\\
    \hline
    \citet{falcon2020data} & Investigate DA techniques for video QA including mirroring and horizontal flipping.\\
    \bottomrule
    \end{tabular}
    \caption{Additional DA works for multimodal tasks, along with a brief description of each.}
    \textbf{\label{table:additional_DA_works_multimodal}}
    \vspace*{5.1in}
    \vspace{-5ex}
\end{table*}

\end{document}